\documentclass[runningheads]{llncs}

\usepackage{accv}

\usepackage{accvabbrv}

\usepackage{graphicx}
\usepackage{booktabs}
\usepackage{multirow}
\usepackage{colortbl}
\usepackage{amsmath}
\usepackage{amssymb}
\usepackage{algorithm}
\usepackage{algpseudocode}

\usepackage[accsupp]{axessibility}  % Improves PDF readability for those with disabilities.

\usepackage{hyperref}

\newcommand{\drop}[1]{($\downarrow$#1)}

\definecolor{bestcell}{HTML}{DCD9FF}

\begin{document}

\title{InfraQR: Edge-Placed QR-Inspired Structured Patch Attacks on Infrared Vision-Language Models}

\titlerunning{InfraQR}

\author{Xin Li\inst{1} \and
Jiaju Han\inst{1,3} \and
Ma Yaqi\inst{2} \and
Chengyin Hu\inst{1} \and
Yingying Zhao\inst{1} \and
Jiahuan Long\inst{4} \and
Fengyu Zhang\inst{1} \and
Yahui Chai\inst{1}}

\authorrunning{X.~Li et al.}

\institute{China University of Petroleum-Beijing at Karamay, Karamay, Xinjiang, China
\and Guizhou University, Guiyang, China
\and Shenzhen Research Institute of Big Data, Shenzhen, China
\and Shanghai Jiao Tong University, Shanghai, China}

\maketitle

\begin{abstract}
  Infrared vision-language models are increasingly used for perception under
  low-light and adverse visual conditions, yet their robustness to localized
  structured perturbations remains underexplored. Existing infrared adversarial
  studies mainly focus on object detectors, leaving the security of infrared
  vision-language models less systematically examined. We present InfraQR, a
  QR-inspired structured patch attack for infrared vision-language models. Unlike
  localized attacks that attach perturbations to the target object, InfraQR places
  a compact structured patch along image boundaries and optimizes learnable grid
  cells through surrogate CLIP-style encoders. The resulting patch has a
  near-binary structured appearance, but is not required to be a valid or
  machine-readable QR code. We evaluate InfraQR on infrared classification,
  caption transfer, and question--answer-aware visual question answering (VQA) tasks. On a 300-image
  infrared benchmark, InfraQR sharply reduces the accuracy of multiple
  CLIP-style classifiers, including reducing OpenAI CLIP accuracy from
  98.67\% to 0.70\%. The generated adversarial images also transfer to
  black-box captioning and VQA models, causing semantic degradation in captions
  and more error-prone answers under GPT-5.4-based evaluation. These results
  show that infrared vision-language models remain vulnerable to structured
  edge-placed perturbations, motivating further study of cross-task robustness
  beyond direct object occlusion.

\end{abstract}

\section{Introduction}
\label{sec:introduction}
Vision-language models (VLMs) have broadened visual perception by aligning
images with natural-language descriptions, enabling open-vocabulary
recognition, captioning, and visual question answering
\cite{radford2021clip,li2023blip2}. Recent studies show that VLMs remain
vulnerable to adversarial perturbations and transfer-based attacks in visible
image domains \cite{liu2025lighting,xie2025chain}. In contrast, the robustness
of infrared VLMs remains less understood, even though infrared imagery is
important for perception under low-light, nighttime, and adverse visual
conditions.

Existing infrared adversarial studies mainly focus on object detectors and
physical-world perturbations, such as learnable infrared patches and infrared
curves \cite{wei2023physically,hu2024advic}. These works reveal important
vulnerabilities of thermal object detection, but they do not directly address
whether structured perturbations can disrupt infrared vision-language
understanding. This gap is important because errors in infrared multimodal
perception may propagate from visual recognition to language generation and
decision making.

Adversarial patch attacks provide a practical way to study localized
vulnerabilities by modifying only a small image region. Most patch settings
place perturbations on or near the target object, directly overlapping the
visual evidence used for recognition. In this work, we ask a complementary
question: can a structured patch placed away from the main object still disrupt
infrared vision-language understanding? This setting is especially relevant for
digital robustness analysis, where the attacker can control image-level
placement without attaching a physical patch to the object itself.

In this work, we study this question through edge-placed structured
perturbations. We introduce InfraQR, a QR-inspired structured patch attack for
infrared VLMs. InfraQR uses a fixed finder-style layout and learnable grid
cells to form a compact local patch with a recognizable structured appearance.
The learnable cells are optimized in a continuous space with a binary
regularization term, which encourages a near-binary pattern without requiring
the patch to be a valid or scannable QR code. Instead of placing the patch on
the target object, InfraQR searches over candidate edge locations and inserts
the patch along the image boundary. This design tests whether infrared VLMs
can be attacked by peripheral structured signals rather than direct object
occlusion.

We evaluate InfraQR from three complementary perspectives. First, we study
classification attacks on multiple CLIP-style infrared classifiers.
Second, we test whether the generated adversarial images transfer to
black-box captioning models and reduce caption semantic consistency. Third,
we extend the objective to question--answer-aware VQA by constructing
question-conditioned semantic targets without accessing the evaluated VQA
models during optimization. Across these settings, InfraQR consistently produces strong degradation compared
with representative patch baselines, showing that structured edge perturbations
can affect not only surrogate classification models but also downstream
language-generation behavior. Our contributions are summarized as follows:
\begin{itemize}
    \item We propose InfraQR, a QR-inspired structured patch attack for infrared
    VLMs that combines finder-style anchors, learnable grid cells, binary
    regularization, and edge-based placement.
    \item We show that localized structured patches do not need to be placed on
    the target object to be effective; edge-placed InfraQR perturbations can
    strongly degrade infrared vision-language representations without directly
    occluding the main visual evidence.
    \item We provide a multi-task evaluation covering infrared classification,
    caption transfer, and question--answer-aware VQA, together with ablations
    on key optimization and patch-design factors.
\end{itemize}

\section{Related Work}
\label{sec:related_work}

\subsection{Adversarial Examples and Localized Patch Attacks}

Adversarial examples show that deep networks can be misled by carefully
constructed perturbations~\cite{szegedy2014intriguing,goodfellow2015explaining}.
Localized attacks further restrict the
modified region to a visible patch. Adversarial Patch introduced universal and
targetable local patterns~\cite{brown2017adversarialpatch}, while LaVAN
showed that visible localized noise can fool classifiers even without covering
the main object~\cite{karmon2018lavan}. Physical-world attacks extend this
idea to road signs, detectors, and person detection through robust or
object-attached patterns~\cite{brown2017adversarialpatch,karmon2018lavan}.
Unlike these object-centered settings, InfraQR studies a digital edge-placed
structured patch for infrared vision-language models.

\subsection{Vision-Language Models and Multimodal Robustness}

Vision-language pretraining has become a foundation for open-vocabulary
recognition and multimodal generation. CLIP aligns images and text through
contrastive learning~\cite{radford2021clip}, and later works scale or improve
image-text representation learning~\cite{cherti2023openclip,gadre2023datacomp,xu2023metaclip,sun2023evaclip}.
Large multimodal models connect visual encoders with language models for
captioning, VQA, and instruction following~\cite{li2023blip2,dai2023instructblip,liu2024llava15,awadalla2023openflamingo}.
Recent studies have increasingly examined adversarial robustness in
vision-language and multimodal systems. Early work attacks vision-language
pretraining models by exploiting image-text interactions and improving
black-box transferability across VLP models~\cite{zhang2022coattack,lu2023sga}.
More recent work further scales this direction through self-supervised
large-scale VLM attacks, transfer-chain optimization, global-local
transformations, and universal perturbations against VLP models~
\cite{zhang2025anyattack,xie2025chain,liu2025gleam,fang2025oneperturbation,li2026transform}.
Other studies expose vulnerabilities of LVLMs to scene-coherent typographic
attacks, illumination transformations, and open-ended VQA attacks~
\cite{cao2025scenetap,liu2025lighting,hu2026omniattack}. Most of these studies
focus on RGB images, full-image perturbations, typographic cues, or standard
multimodal benchmarks. InfraQR instead investigates infrared imagery and asks
whether a structured local patch optimized through surrogate CLIP-style
encoders can transfer to black-box captioning and VQA models.

\subsection{Infrared Adversarial Attacks}

Infrared perception is important for low-light, nighttime, and adverse-weather
scenarios, but its different imaging mechanism changes the form of adversarial
manipulation. Prior infrared attacks mainly focus on detection and physical
settings. HOTCOLD Block uses hot and cold materials to fool thermal detectors~
\cite{wei2023hotcold}; physically adversarial infrared patches optimize
learnable shapes and locations~\cite{wei2023physically}; AdvIB and AdvGrid
study multi-view black-box infrared attacks~\cite{hu2023advib,tiliwalidi2024advgrid};
and unified adversarial patches investigate cross-modal visible-infrared
attacks~\cite{wei2023unified}. AdvIC and other infrared patch-generation
methods further study transferability in thermal imagery~
\cite{hu2024advic}. These works demonstrate infrared
vulnerability, but they mainly target detectors and often attach perturbations
to target objects. Compared with recent VLM attacks that study RGB
typographic or illumination perturbations~\cite{cao2025scenetap,liu2025lighting},
InfraQR focuses on infrared imagery and edge-placed structured patches. This
setting differs from prior infrared attacks on detectors and from RGB VLM
attacks, allowing us to evaluate whether infrared vision-language models are
vulnerable to localized peripheral structured signals.

% ===============================================================
% Method
% ===============================================================
\section{Method}
\label{sec:method}

\subsection{Overview and Threat Model}
\label{sec:threat_model}

\begin{figure*}[t]
\centering
\includegraphics[width=0.98\textwidth]{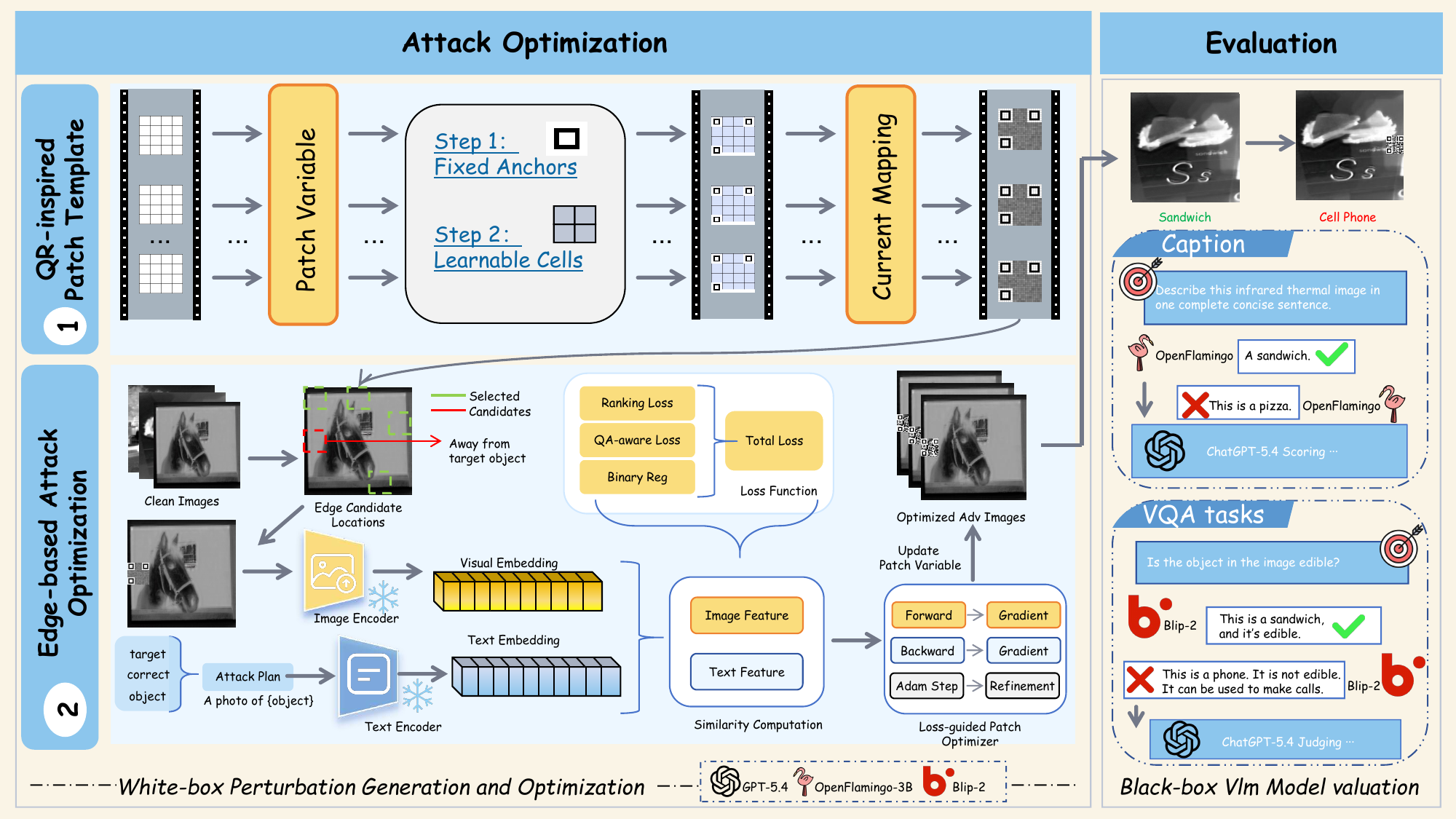}
\caption{Overall framework of InfraQR.}
\label{fig:framework}
\end{figure*}

An overview of the InfraQR framework is provided in
Fig.~\ref{fig:framework}. We propose InfraQR, a QR-inspired structured adversarial patch
framework for infrared vision-language models. InfraQR combines a
partially fixed spatial layout with continuously optimized patch values.
This design preserves a recognizable grid-based structure while retaining
sufficient optimization freedom to manipulate vision-language
representations. The framework consists of three main components:
a structured patch parameterization, an edge-based placement strategy,
and task-dependent semantic attack objectives.

We consider a per-instance localized attack setting. Let
$x \in [0,1]^{H \times W \times 3}$ denote an infrared image after
three-channel input conversion, and let $P$ denote a compact adversarial
pattern defined on a logical grid. We use $\Pi_l(P)$ to represent the
operation that resizes the pattern to the desired patch resolution and
places it at location $l$ on an image-sized canvas. Let $M_l$ be the
corresponding binary spatial mask. The adversarial image is constructed as
\begin{equation}
x^{\mathrm{adv}}
=
(1-M_l)\odot x
+
M_l\odot \Pi_l(P),
\label{eq:patch_composition}
\end{equation}
where $\odot$ denotes element-wise multiplication. In our implementation,
the patch location is selected from a predefined set of boundary
candidates, thereby preserving most of the original image content while
restricting the perturbation to a compact local region.

For the classification-guided setting, the attacker has white-box access
to a surrogate CLIP-style model \cite{radford2021clip}, including its image
encoder $f_I$, text encoder $f_T$, and gradients. Given a set of class
prompts $\mathcal{T}=\{t_k\}_{k=1}^{K}$, we define the normalized
vision-language similarity for class $k$ as
\begin{equation}
s_k(x)
=
\left\langle
\frac{f_I(x)}{\lVert f_I(x)\rVert_2},
\frac{f_T(t_k)}{\lVert f_T(t_k)\rVert_2}
\right\rangle .
\label{eq:vl_similarity}
\end{equation}
The predicted class is
\begin{equation}
\hat{y}(x)
=
\underset{k \in \{1,\ldots,K\}}{\arg\max}\; s_k(x).
\label{eq:clip_prediction}
\end{equation}
Given the ground-truth class $y$, an attack is successful when
\begin{equation}
\hat{y}(x^{\mathrm{adv}}) \neq y.
\label{eq:attack_success}
\end{equation}

The downstream captioning models are not accessed during patch
optimization. Classification-guided adversarial images are directly
transferred to these models to evaluate cross-model and cross-task
generalization. Therefore, the captioning models are treated as black-box
transfer targets.

We additionally consider a question--answer-aware setting for visual
question answering. In this setting, the attacker has access to the
question and its reference answer and uses them to construct a semantic
text target for the surrogate vision-language encoder. However, the
evaluated VQA models and their gradients remain inaccessible during patch
optimization. This setting measures whether a patch optimized in the
surrogate embedding space can transfer to unseen generative VQA models.

\subsection{QR-Inspired Structured Patch Parameterization}
\label{sec:patch_parameterization}

InfraQR parameterizes the adversarial pattern on a compact logical grid
rather than directly optimizing every pixel of the final image-space
patch. Specifically, we use a grid resolution of $G \times G$, where
$G=21$. Three $7 \times 7$ finder-style anchor templates are placed at
the top-left, top-right, and bottom-left corners of the grid. These
anchors remain fixed throughout optimization, while all remaining cells
are treated as learnable variables.

Let
$F \in \{0,1\}^{G \times G}$
denote a binary mask of the fixed anchor cells, where
$F_i=1$ indicates a fixed cell. We use
$A \in \{0,1\}^{G \times G}$
to store the predefined anchor values. The set of learnable cells is
defined as
\begin{equation}
\Omega
=
\left\{
i \mid F_i=0
\right\}.
\label{eq:learnable_region}
\end{equation}

To retain differentiability, the free cells are represented by
continuous logits
$Z \in \mathbb{R}^{G \times G}$.
A temperature-controlled sigmoid maps these logits to the valid intensity
range:
\begin{equation}
S
=
\sigma(\tau Z),
\label{eq:soft_patch_values}
\end{equation}
where $\tau$ controls the sharpness of the mapping. The complete logical
patch is then constructed as
\begin{equation}
P^{\mathrm{log}}
=
F\odot A
+
(1-F)\odot S.
\label{eq:structured_logical_patch}
\end{equation}
Consequently, gradients update only the free region, whereas the three
finder-style anchors preserve their predefined structure.

In our implementation, the learnable logits are initialized as
\begin{equation}
Z_i
\sim
\mathcal{N}(0,0.01^2),
\qquad i\in\Omega,
\label{eq:patch_initialization}
\end{equation}
and the sigmoid temperature is set to $\tau=15$. The small initialization
keeps the initial free-cell intensities close to the middle of the valid
range, avoiding premature sigmoid saturation and allowing stable
gradient-based optimization.

The logical pattern is resized to the required image-space patch
resolution using nearest-neighbor interpolation. This operation preserves
the grid boundaries and the piecewise-constant appearance of individual
cells. For models requiring a three-channel input, the resulting
single-channel intensity pattern is replicated across the three input
channels before being placed on the infrared image.

Although the free cells remain continuous during optimization, we
encourage them to approach the endpoints of the intensity interval using
a binary regularization term:
\begin{equation}
\mathcal{L}_{\mathrm{bin}}
=
\frac{1}{|\Omega|}
\sum_{i\in\Omega}
P^{\mathrm{log}}_i
\left(
1-P^{\mathrm{log}}_i
\right).
\label{eq:binary_regularization}
\end{equation}
Minimizing \cref{eq:binary_regularization} drives the learnable values
toward $0$ or $1$ while preserving a differentiable optimization path.
Importantly, the regularization is evaluated only over $\Omega$ because
the anchor cells are already fixed and should not contribute to the
optimization objective.

This parameterization produces a binary-regularized soft patch with a
stable QR-inspired layout. The pattern borrows the grid organization and
finder-style visual anchors of QR codes, but it is not constrained by QR
encoding, error-correction, or machine-readability requirements.

\subsection{Classification-Guided Attack Objective}
\label{sec:classification_objective}

Given an adversarial image $x^{\mathrm{adv}}$, we compute its normalized
similarity to each class prompt using \cref{eq:vl_similarity}. Let
\begin{equation}
\tilde{s}_y^{\mathrm{adv}}
=
\operatorname{sim}(x^{\mathrm{adv}}, y)
\label{eq:ground_truth_similarity}
\end{equation}
denote the normalized similarity score of the ground-truth class $y$, where $\operatorname{sim}(x,c)$ is the normalized image--text similarity between image $x$ and the class prompt of category $c$. We
further define the strongest competing-class similarity as
\begin{equation}
s_{\mathrm{comp}}^{\mathrm{adv}}
=
\max_{c\neq y}
\operatorname{sim}(x^{\mathrm{adv}}, c).
\label{eq:competing_similarity}
\end{equation}

The corresponding semantic ranking gap is
\begin{equation}
g(x^{\mathrm{adv}})
=
\tilde{s}_y^{\mathrm{adv}}
-
s_{\mathrm{comp}}^{\mathrm{adv}}.
\label{eq:semantic_gap}
\end{equation}
A positive gap indicates that the ground-truth class remains ranked
above all competing classes, whereas a negative gap indicates that at
least one competing class has surpassed it.

We optimize a ranking loss:
\begin{equation}
\mathcal{L}_{\mathrm{cls}}
=
\max
\left(
0,\,
\tilde{s}_y^{\mathrm{adv}}
-
s_{\mathrm{comp}}^{\mathrm{adv}}
\right).
\label{eq:classification_attack_loss}
\end{equation}
This objective penalizes the patch while the ground-truth similarity
exceeds the competing similarity. The attack is counted as successful
when the final prediction satisfies the condition in \cref{eq:attack_success}.

The complete optimization objective combines the classification-guided
loss with the binary regularization term introduced in
\cref{eq:binary_regularization}:
\begin{equation}
\mathcal{L}_{\mathrm{total}}
=
\mathcal{L}_{\mathrm{cls}}
+
\lambda_{\mathrm{bin}}
\mathcal{L}_{\mathrm{bin}},
\label{eq:classification_total_loss}
\end{equation}
where $\lambda_{\mathrm{bin}}$ controls the trade-off between attack
effectiveness and near-binary patch values. Gradients of
\cref{eq:classification_total_loss} are propagated only to the learnable
logits in $\Omega$; the finder-style anchors remain fixed throughout
optimization.

Starting from the initialization in
\cref{eq:patch_initialization}, the learnable logits are iteratively
updated using a gradient-based optimizer. At every iteration, the current
logical pattern is reconstructed using
\cref{eq:structured_logical_patch}, resized to the selected image-space
resolution, and composed with the input image according to
\cref{eq:patch_composition}.

When optimization terminates, whether due to satisfaction of the success
criterion or exhaustion of the iteration budget, we reconstruct the patch
from the final logits and perform a separate forward pass on the resulting
adversarial image. Attack success is then determined from this final
prediction using \cref{eq:attack_success}, rather than being inferred
solely from the training loss or an intermediate optimization state.

\subsection{Edge-Based Placement Search}
\label{sec:placement_search}

The effectiveness of a localized patch can vary substantially with its
spatial position because different image regions contribute unequally to
the surrogate representation. Jointly optimizing a continuous location
with the patch values would introduce additional complexity and could
make comparisons between different patch designs less controlled. We
therefore adopt a lightweight discrete placement search before optimizing
the patch content.

Let $\mathcal{L}$ denote a predefined set of boundary locations. We
define the candidate set by uniformly sampling five locations on each image
edge, including the top, bottom, left, and right boundaries. This produces
20 edge-indexed candidates in total. Because adjacent edges share corner
coordinates, these candidates correspond to 16 unique spatial locations.
For a patch of spatial size $p \times p$, all candidates are constrained
to keep the complete patch within the image boundary.

To estimate the sensitivity of each location without favoring a
particular attack pattern, we use a constant gray probe patch
\begin{equation}
P^{\mathrm{probe}}
=
\rho_{\mathrm{probe}}\mathbf{1}_{p\times p},
\label{eq:probe_patch}
\end{equation}
where $\rho_{\mathrm{probe}}=0.5$ denotes the fixed gray intensity used for
the probe patch. For each candidate $l\in\mathcal{L}$, the probe is inserted into the
input image to produce $x_l^{\mathrm{probe}}$. We then evaluate the
ground-truth similarity using the surrogate model and select
\begin{equation}
l^{\star}
=
\underset{l\in\mathcal{L}}{\arg\min}\;
s_y\!\left(x_l^{\mathrm{probe}}\right).
\label{eq:placement_selection}
\end{equation}
Thus, the selected location is the boundary position at which a neutral
local replacement causes the largest reduction in the ground-truth
semantic similarity.

The placement search requires only forward evaluations and is performed
once for each image and surrogate model. After obtaining $l^{\star}$,
the location remains fixed throughout patch optimization. For controlled comparison, the same selected location is also used by
HCB~\cite{wei2023hotcold}, AdvIC~\cite{hu2024advic},
AdvGrid~\cite{tiliwalidi2024advgrid}, and InfraQR, ensuring that performance
differences are primarily attributable to their patch structures and optimization
strategies rather than to different spatial placements.
Algorithm~\ref{alg:infraq_classification} summarizes the overall InfraQR
procedure, including edge-based placement search, structured patch
construction, and iterative patch optimization.

\begin{algorithm}[t]
\caption{Pseudocode of InfraQR}
\label{alg:infraq_classification}
\begin{algorithmic}[1]
\Require Infrared image $x$, ground-truth label $y$, surrogate encoders
$f_I$ and $f_T$, class prompts $\mathcal{T}$, boundary candidates
$\mathcal{L}$, iteration budget $T$, learning rate $\eta$, temperature
$\tau$, and regularization weight $\lambda_{\mathrm{bin}}$
\Ensure Final adversarial image $x^{\mathrm{adv}}$ and success indicator

\State Encode and normalize all class prompts using $f_T$
\State Construct the constant probe patch
$P^{\mathrm{probe}} \leftarrow \rho_{\mathrm{probe}}\mathbf{1}_{p\times p}$

\For{each candidate location $l\in\mathcal{L}$}
    \State Compose $x_l^{\mathrm{probe}}$ by placing
    $P^{\mathrm{probe}}$ at $l$
    \State Compute the ground-truth similarity
    $s_y(x_l^{\mathrm{probe}})$
\EndFor
\State Select
$l^{\star}\gets
\arg\min_{l\in\mathcal{L}}s_y(x_l^{\mathrm{probe}})$

\State Construct the fixed masks $F$ and anchor values
$A$
\State Initialize
$Z_i\sim\mathcal{N}(0,0.01^2)$ for $i\in\Omega$
\State Initialize the Adam optimizer with learning rate $\eta$

\For{$t=1$ to $T$}
    \State $S\gets\sigma(\tau Z)$
    \State $P^{\mathrm{log}}\gets
    F\odot A
    +(1-F)\odot S$
    \State Resize $P^{\mathrm{log}}$ and place it at
    $l^{\star}$ to obtain $x_t^{\mathrm{adv}}$
    \State Compute similarities
    $\{s_k(x_t^{\mathrm{adv}})\}_{k=1}^{K}$

    \If{$\hat{y}(x_t^{\mathrm{adv}})\neq y$}
        \State \textbf{break}
    \EndIf

    \State $s_{\mathrm{comp}}\gets
    \max_{k\neq y}s_k(x_t^{\mathrm{adv}})$
    \State $\mathcal{L}_{\mathrm{cls}}\gets
    \max(0,s_y-s_{\mathrm{comp}})$
    \State Compute $\mathcal{L}_{\mathrm{bin}}$ only over $\Omega$
    \State $\mathcal{L}_{\mathrm{total}}\gets
    \mathcal{L}_{\mathrm{cls}}
    +\lambda_{\mathrm{bin}}\mathcal{L}_{\mathrm{bin}}$
    \State Update $Z$ using Adam on
    $\nabla_{Z}\mathcal{L}_{\mathrm{total}}$
\EndFor

\State Reconstruct the patch using the final logits $Z$
\State Perform an independent final forward pass to obtain
$x^{\mathrm{adv}}$ and $\hat{y}(x^{\mathrm{adv}})$
\State \Return $x^{\mathrm{adv}}$ and
$\mathbb{I}\!\left[\hat{y}(x^{\mathrm{adv}})\neq y\right]$
\end{algorithmic}
\end{algorithm}

\subsection{Question--Answer-Aware Extension}
\label{sec:qa_extension}

For visual question answering, the attack objective should depend on both the
question and the reference answer. We extend InfraQR to question--answer-conditioned
semantic targets. Let $q$ denote a question, $r$ its question type, and $a^{+}$
the canonical reference answer. A template function $\phi(\cdot)$ converts the
triplet $(q,a,r)$ into a text target:
\begin{equation}
t(q,a,r)
=
\phi(q,a,r).
\label{eq:qa_text_target}
\end{equation}
The correct target is written as
\begin{equation}
t^{+}
=
t(q,a^{+},r).
\label{eq:qa_positive_target}
\end{equation}
The template is selected according to the question type to preserve the
required semantics. This allows the CLIP-style surrogate encoder to provide
gradients for a VQA-oriented objective without accessing the evaluated VQA model.

For objective question types with a reliable answer space, we construct a
candidate answer set $\mathcal{A}(q,r)$. The negative candidate set is
\begin{equation}
\mathcal{A}^{-}(q,r)
=
\mathcal{A}(q,r)\setminus\{a^{+}\}.
\label{eq:qa_negative_set}
\end{equation}
This set is used only when plausible incorrect answers can be generated
reliably. We do not assume that arbitrary open-ended questions admit reliable
automatically generated negative answers.

Given an adversarial image $x^{\mathrm{adv}}$, the similarity to the correct
question--answer target is
\begin{equation}
s^{+}(x^{\mathrm{adv}})
=
\left\langle
\frac{f_I(x^{\mathrm{adv}})}{\lVert f_I(x^{\mathrm{adv}})\rVert_2},
\frac{f_T(t^{+})}{\lVert f_T(t^{+})\rVert_2}
\right\rangle .
\label{eq:qa_positive_similarity}
\end{equation}
When $\mathcal{A}^{-}(q,r)$ is non-empty, we further define the strongest
incorrect target in the surrogate embedding space as
\begin{equation}
s^{-}(x^{\mathrm{adv}})
=
\max_{a\in\mathcal{A}^{-}(q,r)}
\left\langle
\frac{f_I(x^{\mathrm{adv}})}{\lVert f_I(x^{\mathrm{adv}})\rVert_2},
\frac{f_T(t(q,a,r))}{\lVert f_T(t(q,a,r))\rVert_2}
\right\rangle .
\label{eq:qa_negative_similarity}
\end{equation}
The QA-aware ranking objective is
\begin{equation}
\mathcal{L}_{\mathrm{qa}}
=
s^{+}(x^{\mathrm{adv}})
-
s^{-}(x^{\mathrm{adv}}).
\label{eq:qa_ranking_loss}
\end{equation}
Minimizing Eq.~\eqref{eq:qa_ranking_loss} suppresses the surrogate similarity
to the correct question--answer target while promoting the most competitive
incorrect target. If no reliable negative candidate is available, the objective
reduces to suppressing the correct target similarity:
\begin{equation}
\mathcal{L}_{\mathrm{qa}}
=
s^{+}(x^{\mathrm{adv}}).
\label{eq:qa_positive_only_loss}
\end{equation}

The final QA-aware InfraQR objective combines the semantic objective with the
same binary regularization used in the classification-guided setting:
\begin{equation}
\mathcal{L}_{\mathrm{qa,total}}
=
\mathcal{L}_{\mathrm{qa}}
+
\lambda_{\mathrm{bin}}\mathcal{L}_{\mathrm{bin}}.
\label{eq:qa_total_loss}
\end{equation}
As before, binary regularization is computed only over the learnable cells
$\Omega$, while the finder-style anchors remain fixed. The evaluated VQA
models are not queried during optimization, and their gradients are never used.

\section{Experiments}
\label{sec:experiments}

\subsection{Experimental Setup}
\label{sec:experimental_setup}

\paragraph{Datasets.}
For classification and caption-transfer evaluation, we construct a 300-image
infrared benchmark from Infrared-Image-Instruct-12K.
The benchmark contains 30 object categories with 10 images per category. We
first use a vision-language model to select, for each target category,
images whose visual content best matches the category description, and then
manually verify the selected images. All images are converted to three-channel inputs and resized to
$224\times224$.

For VQA, we build a curated objective subset from an infrared VQA manifest
with 1430 candidate questions. We retain only questions with reliable short
answers, including yes/no, counting, spatial-location, and object-category
questions. The final subset contains 295 image-question pairs, with 59
validation samples and 236 test samples. The selected questions include 117
yes/no, 74 counting, 61 object-category, and 43 spatial-location questions.
The main VQA results are reported on the test split.

\paragraph{Models.}
For classification-guided attacks, we use four CLIP-style surrogate encoders:
OpenAI CLIP ViT-L/14 \cite{radford2021clip}, MetaCLIP ViT-L/14
\cite{xu2023metaclip}, EVA-CLIP ViT-G/14 \cite{sun2023evaclip}, and
OpenCLIP/DataComp ViT-B/16 \cite{cherti2023openclip,gadre2023datacomp}. Each
class is represented by the prompt template ``a thermal photo of a
\{category\}''. For caption transfer, we evaluate the generated adversarial
images on BLIP-2 \cite{li2023blip2}, InstructBLIP
\cite{dai2023instructblip}, LLaVA-1.5 \cite{liu2024llava15}, LLaVA-1.6~\cite{liu2024llavanext}, and
OpenFlamingo \cite{awadalla2023openflamingo}. For VQA, OpenAI CLIP ViT-L/14
and EVA-CLIP ViT-G/14 are used as QA-aware surrogate encoders, and the
resulting attacks are transferred to black-box generative VQA models.

\paragraph{Baselines and attack settings.}
We compare InfraQR with three infrared physical-attack-inspired baselines:
HCB \cite{wei2023hotcold}, AdvIC \cite{hu2024advic}, and AdvGrid
\cite{tiliwalidi2024advgrid}. All methods use the same image preprocessing,
candidate boundary locations, and selected patch placement for each image and
surrogate encoder. Unless otherwise specified, the patch size is $37\times37$
on a $224\times224$ image. InfraQR uses a $21\times21$ logical grid with three
fixed $7\times7$ finder-style anchors. The learnable logits are optimized for
1000 iterations using Adam with learning rate $0.03$, sigmoid temperature
$\tau=15$, initialization standard deviation $0.01$, and binary regularization
weight $\lambda_{\mathrm{bin}}=0.01$. All experiments use random seed 42.

\paragraph{Evaluation.}
For classification, we report clean accuracy, post-attack classification
accuracy, and absolute performance degradation measured by accuracy drop. Lower
post-attack accuracy and larger accuracy drop indicate stronger attack
performance. When reporting attack success rate (ASR), we compute it on
clean-correct samples. For caption transfer, the clean caption from the same
target model serves as a paired semantic reference, and we report semantic
consistency after attack; lower consistency indicates stronger caption
degradation. For VQA, we report answer correctness after attack, where lower
correctness indicates stronger degradation. Captioning and VQA models are
treated as black-box targets and are not queried during attack optimization.
For caption transfer, GPT-5.4~\cite{openai2026gpt54} scores the semantic
consistency between each attacked caption and the paired clean caption. For VQA, GPT-5.4 serves as the
official judge of answer correctness against the accepted reference answers.

\subsection{Classification Attack Results}
\label{sec:classification_results}

\begin{table*}[t]
\centering
\caption{Classification results under different adversarial attacks.}
\label{tab:classification_results}
\setlength{\tabcolsep}{4.5pt}
\renewcommand{\arraystretch}{1.08}
\resizebox{0.98\textwidth}{!}{
\begin{tabular}{ccccccccc}
\toprule
\multirow{2}{*}{\textbf{Method}}
& \multicolumn{2}{c}{\textbf{OpenCLIP ViT-B/16}}
& \multicolumn{2}{c}{\textbf{MetaCLIP ViT-L/14}}
& \multicolumn{2}{c}{\textbf{EVA-CLIP ViT-G/14}}
& \multicolumn{2}{c}{\textbf{OpenAI CLIP ViT-L/14}} \\
\cmidrule(lr){2-3} \cmidrule(lr){4-5} \cmidrule(lr){6-7} \cmidrule(lr){8-9}
& \textbf{Acc} & \textbf{Drop}
& \textbf{Acc} & \textbf{Drop}
& \textbf{Acc} & \textbf{Drop}
& \textbf{Acc} & \textbf{Drop} \\
\midrule
Clean   & 96.00 & --    & 96.67 & --    & 98.33 & --    & 98.67 & -- \\
HCB     & 91.49 & 4.51  & 94.95 & 1.72  & 97.65 & 0.68  & 96.98 & 1.69 \\
AdvIC   & 80.38 & 15.62 & 91.15 & 5.52  & 95.96 & 2.37  & 86.85 & 11.82 \\
AdvGrid & 85.58 & 10.42 & 93.22 & 3.45  & 97.31 & 1.02  & 91.91 & 6.76 \\
InfraQR & \cellcolor{bestcell}\textbf{3.64} & \cellcolor{bestcell}\textbf{92.36}
        & \cellcolor{bestcell}\textbf{10.81} & \cellcolor{bestcell}\textbf{85.86}
        & \cellcolor{bestcell}\textbf{62.40} & \cellcolor{bestcell}\textbf{35.93}
        & \cellcolor{bestcell}\textbf{0.70} & \cellcolor{bestcell}\textbf{97.97} \\
\bottomrule
\end{tabular}}
\end{table*}

Table~\ref{tab:classification_results} reports the direct classification
attack results on four CLIP-style infrared surrogate encoders. The clean
accuracies are high across all encoders, ranging from 96.00\% to 98.67\%,
indicating that the curated benchmark provides reliable category supervision
before attack. Among the three baselines, AdvIC is generally the strongest,
but its largest accuracy drop is 15.62 percentage points on OpenCLIP and
11.82 percentage points on OpenAI CLIP. HCB and AdvGrid produce smaller
degradations in most settings.

\begin{figure*}[t]
\centering
\includegraphics[width=0.98\textwidth]{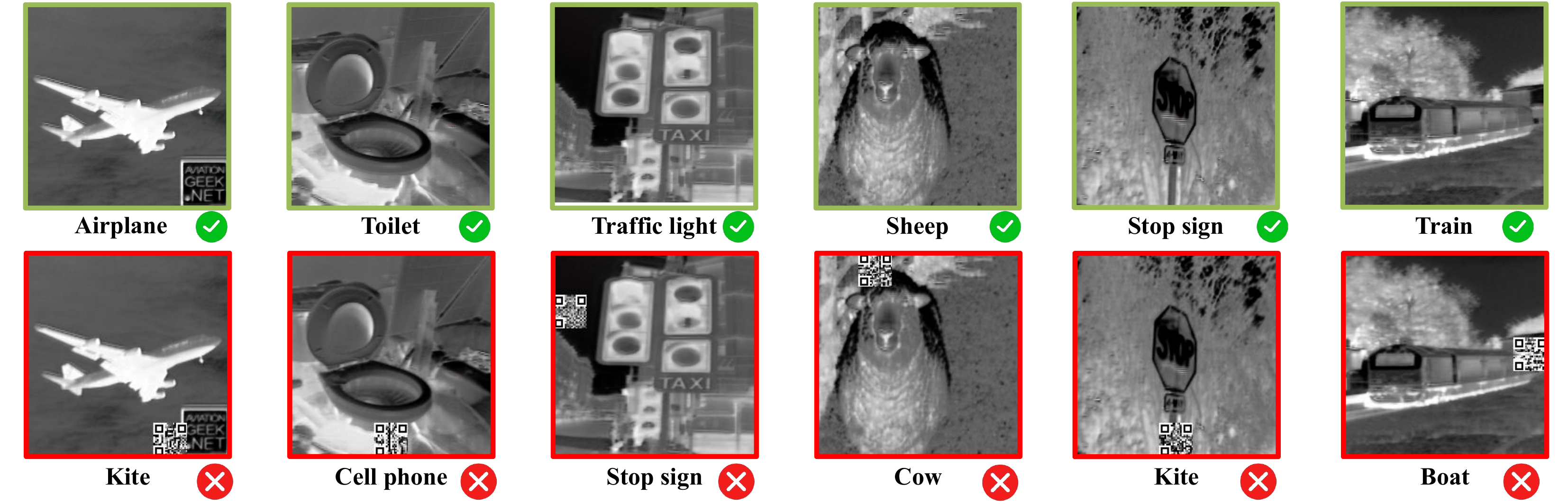}
\caption{Qualitative classification examples under different adversarial attacks.}
\label{fig:qual_cls}
\end{figure*}

InfraQR consistently achieves the largest degradation across all four
surrogate encoders. On OpenAI CLIP ViT-L/14, the accuracy decreases from
98.67\% to 0.70\%, corresponding to a 97.97-point drop. Similar trends are
observed on OpenCLIP and MetaCLIP, where InfraQR reduces the accuracy to
3.64\% and 10.81\%, respectively. EVA-CLIP remains more robust than the
other encoders under all attacks, yet InfraQR still reduces its accuracy by
35.93 points, substantially exceeding the degradation caused by the
baselines. These results show that the QR-inspired structured patch can
strongly disrupt CLIP-style infrared representations under the same localized
placement protocol. Representative qualitative examples are shown in
Fig.~\ref{fig:qual_cls}.

\subsection{Cross-Task Caption Transfer}
\label{sec:caption_transfer}

\begin{table*}[t]
\centering
\caption{Image captioning results under different adversarial attacks; parentheses denote absolute drops from clean samples.}
\label{tab:caption_transfer}
\footnotesize
\setlength{\tabcolsep}{4.0pt}
\renewcommand{\arraystretch}{1.10}
\resizebox{0.98\textwidth}{!}{
\begin{tabular}{c l c c c c c c}
\toprule
\textbf{Image Encoder} & \textbf{Models} & \textbf{Params}
& \textbf{Clean} & \textbf{HCB} & \textbf{AdvIC} & \textbf{AdvGrid} & \textbf{InfraQR} \\
\midrule
\multirow{4}{*}{\begin{tabular}{c}OpenAI CLIP\\ViT-L/14\end{tabular}}
& LLaVA-1.5 & 7B
& 100.00 & 85.62\drop{14.38} & 81.97\drop{18.03} & 81.59\drop{18.41} & \cellcolor{bestcell}\textbf{76.96\drop{23.04}} \\
& LLaVA-1.6 & 7B
& 100.00 & 70.27\drop{29.73} & \cellcolor{bestcell}\textbf{66.55\drop{33.45}} & 68.52\drop{31.48} & 66.97\drop{33.03} \\
& OpenFlamingo & 3B
& 100.00 & 78.74\drop{21.26} & 76.53\drop{23.47} & 79.81\drop{20.19} & \cellcolor{bestcell}\textbf{75.30\drop{24.70}} \\
& BLIP-2 (FlanT5XL ViT-L) & 3.4B
& 100.00 & 89.66\drop{10.34} & 88.18\drop{11.82} & 88.48\drop{11.52} & \cellcolor{bestcell}\textbf{83.48\drop{16.52}} \\
\midrule
\multirow{2}{*}{\begin{tabular}{c}EVA-CLIP\\ViT-G/14\end{tabular}}
& BLIP-2 (FlanT5XL) & 4.1B
& 100.00 & 89.76\drop{10.24} & 87.28\drop{12.72} & 88.70\drop{11.30} & \cellcolor{bestcell}\textbf{86.93\drop{13.07}} \\
& InstructBLIP (FlanT5XL) & 4.1B
& 100.00 & 93.13\drop{6.87} & \cellcolor{bestcell}\textbf{91.76\drop{8.24}} & 93.09\drop{6.91} & 92.50\drop{7.50} \\
\bottomrule
\end{tabular}}
\end{table*}

Table~\ref{tab:caption_transfer} reports captioning transfer results for
classification-guided adversarial images on black-box VLMs. Since the
captioning models are not used during patch optimization, this setting measures
both cross-model and cross-task transferability. InfraQR produces the largest
semantic-consistency drop in four of the six transfer settings, including
BLIP-2, OpenFlamingo, and LLaVA-1.5 under the OpenAI CLIP surrogate, as well
as BLIP-2 under the EVA-CLIP surrogate.

The transfer effect varies across target models. On LLaVA-1.6,
AdvIC is slightly stronger than InfraQR, while InfraQR remains close to the best
result. On InstructBLIP with the EVA-CLIP surrogate, AdvIC also gives the
largest drop, and InfraQR ranks second. These cases suggest that caption-level
transfer depends on the target model architecture and its visual encoder.
Nevertheless, compared with HCB and AdvGrid, InfraQR consistently causes a
stronger or competitive degradation, showing that a structured localized patch
optimized for CLIP-style classification can also disrupt downstream caption
semantics.

\subsection{Question--Answer-Aware VQA Results}
\label{sec:vqa_results}

\begin{table*}[t]
\centering
\caption{VQA results under different adversarial attacks; parentheses denote absolute drops from clean samples.}
\label{tab:vqa_results}
\footnotesize
\setlength{\tabcolsep}{4.0pt}
\renewcommand{\arraystretch}{1.10}
\resizebox{0.98\textwidth}{!}{
\begin{tabular}{c l c c c c c c}
\toprule
\textbf{Image Encoder} & \textbf{Models} & \textbf{Params}
& \textbf{Clean} & \textbf{HCB} & \textbf{AdvIC} & \textbf{AdvGrid} & \textbf{InfraQR} \\
\midrule
\multirow{4}{*}{\begin{tabular}{c}OpenAI CLIP\\ViT-L/14\end{tabular}}
& LLaVA-1.5 & 7B
& 68.90 & 60.00\drop{8.90} & 60.90\drop{8.00} & 59.10\drop{9.80} & \cellcolor{bestcell}\textbf{52.90\drop{16.00}} \\
& LLaVA-1.6 & 7B
& 65.20 & 57.00\drop{8.20} & 58.00\drop{7.20} & 55.00\drop{10.20} & \cellcolor{bestcell}\textbf{48.90\drop{16.30}} \\
& OpenFlamingo & 3B
& 64.80 & 57.00\drop{7.80} & 58.00\drop{6.80} & 55.90\drop{8.90} & \cellcolor{bestcell}\textbf{50.90\drop{13.90}} \\
& BLIP-2 (FlanT5XL ViT-L) & 3.4B
& 60.80 & 55.90\drop{4.90} & 56.90\drop{3.90} & 55.10\drop{5.70} & \cellcolor{bestcell}\textbf{49.70\drop{11.10}} \\
\midrule
\multirow{2}{*}{\begin{tabular}{c}EVA-CLIP\\ViT-G/14\end{tabular}}
& BLIP-2 (FlanT5XL) & 4.1B
& 61.10 & 57.90\drop{3.20} & 59.50\drop{1.60} & 58.70\drop{2.40} & \cellcolor{bestcell}\textbf{45.80\drop{15.30}} \\
& InstructBLIP (FlanT5XL) & 4.1B
& 63.20 & 58.90\drop{4.30} & 59.70\drop{3.50} & 59.50\drop{3.70} & \cellcolor{bestcell}\textbf{46.60\drop{16.60}} \\
\bottomrule
\end{tabular}}
\end{table*}

Table~\ref{tab:vqa_results} reports the QA-aware transfer results on the
curated objective VQA benchmark. In contrast to caption transfer, the attack
objective explicitly uses the question and reference answer to construct
question--answer-conditioned semantic targets, while the evaluated VQA models
remain black-box. The clean VQA accuracies range from 60.8\% to 68.9\%,
showing that the selected questions provide meaningful short-answer
evaluation for infrared VLMs.

\begin{figure*}[t]
\centering
\includegraphics[width=0.98\textwidth]{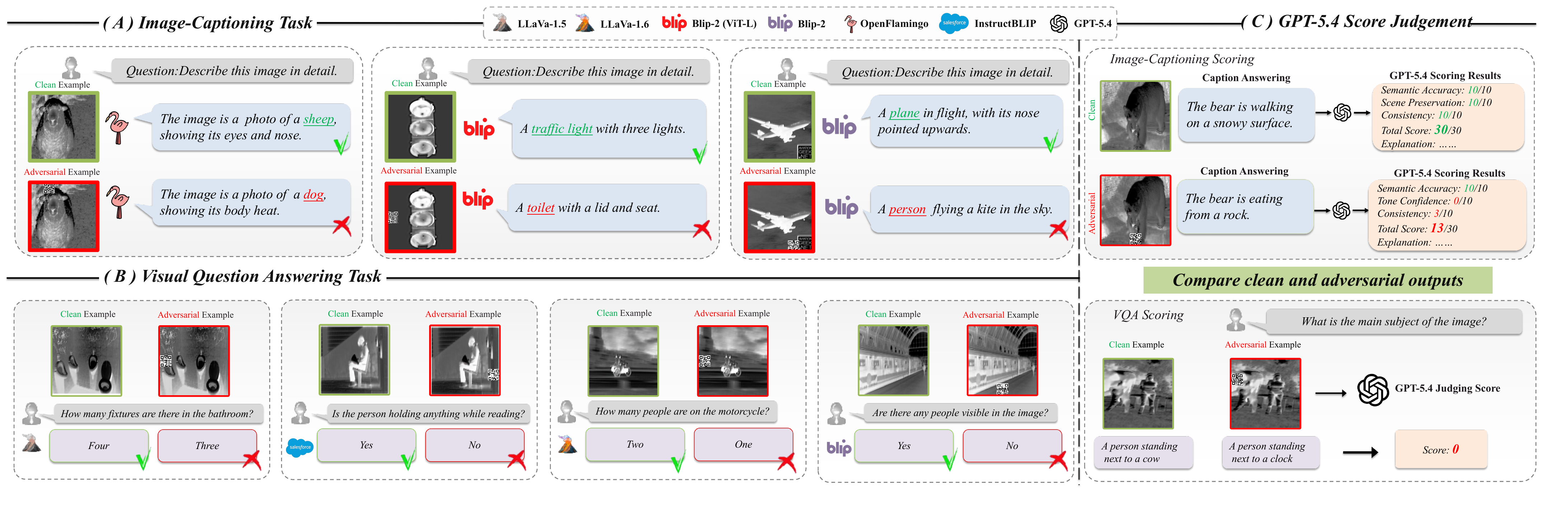}
\caption{Qualitative image-captioning and VQA examples under different adversarial attacks.}
\label{fig:qual_caption_vqa}
\end{figure*}

InfraQR achieves the lowest VQA accuracy in all six transfer settings. Under
the OpenAI CLIP surrogate, InfraQR reduces the accuracy of BLIP-2,
OpenFlamingo, LLaVA-1.5, and LLaVA-1.6 to 49.7\%, 50.9\%, 52.9\%, and
48.9\%, respectively. Under the EVA-CLIP surrogate, InfraQR further reduces
BLIP-2 and InstructBLIP to 45.8\% and 46.6\%. Compared with HCB, AdvIC, and
AdvGrid, the proposed QA-aware objective consistently produces stronger
degradation, indicating that InfraQR can transfer from surrogate
question--answer semantics to black-box VQA behavior.
Fig.~\ref{fig:qual_caption_vqa} provides qualitative examples of the observed
caption and VQA degradations.

\subsection{Ablation and Discussion}
\label{sec:ablation_discussion}

\begin{figure*}[t]
\centering
\includegraphics[width=0.98\textwidth]{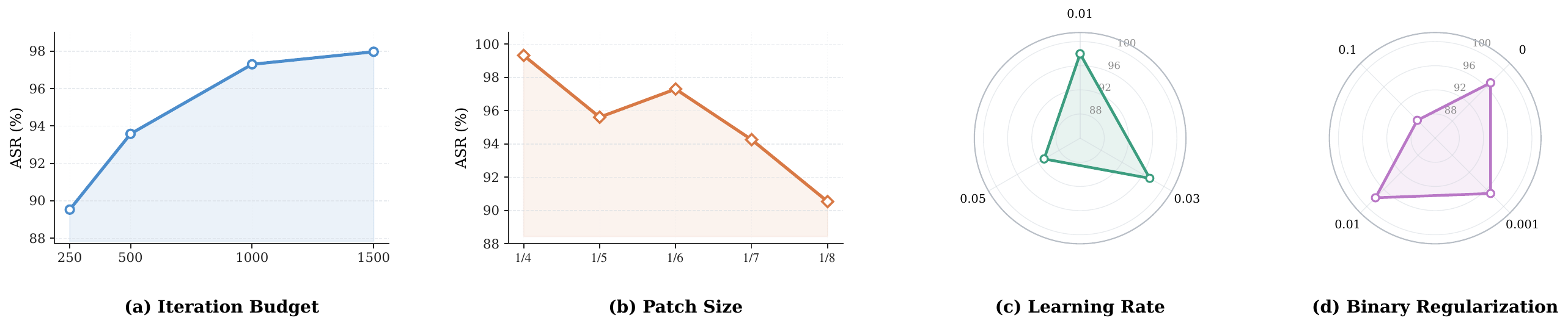}
\caption{Ablation analysis of InfraQR.}
\label{fig:ablation}
\end{figure*}

Fig.~\ref{fig:ablation} summarizes the sensitivity of InfraQR to key
optimization and patch parameters. Increasing the iteration budget improves
ASR from 89.53\% at 250 iterations to 97.97\% at 1500 iterations, with a
smaller gain after 1000 iterations. We therefore use 1000 iterations as the
default setting. The learning-rate ablation shows that overly aggressive
updates hurt stability: 0.01 and 0.03 both achieve high ASR, whereas 0.05
reduces ASR to 90.88\%. Patch size also affects attack capacity. A larger
$1/4$ patch reaches 99.32\% ASR, but occupies more image area; the default
$1/6$ patch keeps the perturbation compact while maintaining strong attack
performance.

The binary-regularization ablation shows that a moderate weight is preferable.
Removing the regularizer still yields high ASR but provides weaker pressure
toward near-binary values, while $\lambda_{\mathrm{bin}}=0.1$ overly restricts
the learnable cells and reduces ASR to 88.18\%. Overall, the default setting
balances attack effectiveness, compact edge placement, and QR-inspired
near-binary structure. InfraQR is not intended to produce a valid or scannable
QR code; its structure provides a local prior while preserving continuous
optimization freedom.

\section{Conclusion}
\label{sec:conclusion}

This study presented InfraQR, an edge-placed QR-inspired structured patch
attack for infrared vision-language models. By combining finder-style anchors,
learnable grid cells, binary regularization, and boundary placement, InfraQR
tests whether structured perturbations can disrupt infrared VLMs without being
placed on the target object. Across classification, caption transfer, and
question--answer-aware VQA, InfraQR degrades CLIP-style infrared recognition
and transfers to black-box generative models. These results indicate that image
boundaries should not be treated as semantically irrelevant regions for
infrared vision-language models. Although the perturbation is spatially
separated from the main object, it can still bias global image-text
representations and affect downstream language outputs. This observation
separates the studied vulnerability from simple object occlusion and highlights
the need to evaluate peripheral structured perturbations when assessing
infrared multimodal robustness. The ablation study further shows
that optimization budget, patch size, learning rate, and binary regularization
jointly affect attack behavior. Our study focuses on digital attacks and does
not claim that the learned pattern is a valid or machine-readable QR code, nor
that it is physically robust to printing, display, sensor noise, or viewpoint
changes. Extending InfraQR to physically realizable infrared settings, together
with defenses against structured peripheral perturbations, is an important
direction for future work.

\bibliographystyle{splncs04}
\bibliography{main,references}
\end{document}